\documentclass[10pt,twocolumn,letterpaper]{article}

\usepackage{iccv}
\usepackage{times}
\usepackage{epsfig}
\usepackage{graphicx}
\usepackage{amsmath}
\usepackage{amssymb}

\usepackage{enumitem}

\usepackage[pagebackref=true,breaklinks=true,letterpaper=true,colorlinks,bookmarks=false,hyperfootnotes=false]{hyperref}

\usepackage{url}
\usepackage{amsfonts}
\usepackage{multirow}
\usepackage[ruled,vlined,linesnumbered,noresetcount]{algorithm2e}
\usepackage{xcolor}
\DeclareMathOperator*{\argmax}{arg\,max}

\iccvfinalcopy 
\usepackage[stable]{footmisc}

\def\iccvPaperID{10277} 

\ificcvfinal\fi

\begin{document}

\title{3D Neural Embedding Likelihood: Probabilistic Inverse Graphics\\for Robust 6D Pose Estimation}

\author{Guangyao Zhou\thanks{Equal contribution} \\
Google DeepMind\\
\texttt{stannis@google.com}
\and
Nishad Gothoskar\footnotemark[1]\\
MIT\\
\texttt{nishad@mit.edu}
\and
Lirui Wang\\
MIT\\
\texttt{liruiw@mit.edu}
\and
Joshua B. Tenenbaum\\
MIT\\
\texttt{jbt@mit.edu}
\and Dan Gutfreund\\
MIT-IBM Watson AI Lab\\
\texttt{dgutfre@us.ibm.com}
\and
Miguel L\'{a}zaro-Gredilla\\
Google DeepMind\\
\texttt{lazarogredilla@google.com}
\and
Dileep George\\
Google DeepMind\\
\texttt{dileepgeorge@google.com}
\and Vikash K. Mansinghka \\
MIT\\
\texttt{vkm@mit.edu}
}

\maketitle
\ificcvfinal\thispagestyle{empty}\fi

\begin{abstract}
	The ability to perceive and understand 3D scenes is crucial for many applications in computer vision and robotics. Inverse graphics is an appealing approach to 3D scene understanding that aims to infer the 3D scene structure from 2D images. In this paper, we introduce probabilistic modeling to the inverse graphics framework to quantify uncertainty and achieve robustness in 6D pose estimation tasks. Specifically, we propose 3D Neural Embedding Likelihood (3DNEL) as a unified probabilistic model over RGB-D images, and develop efficient inference procedures on 3D scene descriptions. 3DNEL effectively combines learned neural embeddings from RGB with depth information to improve robustness in sim-to-real 6D object pose estimation from RGB-D images. Performance on the YCB-Video dataset is on par with state-of-the-art yet is much more robust in challenging regimes. In contrast to discriminative approaches, 3DNEL's probabilistic generative formulation jointly models multiple objects in a scene, quantifies uncertainty in a principled way, and handles object pose tracking under heavy occlusion. Finally, 3DNEL provides a principled framework for incorporating prior knowledge about the scene and objects, which allows natural extension to additional tasks like camera pose tracking from video.
\end{abstract}

\section{Introduction}\label{sec:intro}
3D scene understanding is a fundamental problem in computer vision and robotics with numerous applications, including object recognition \cite{xiang2014beyond}, robotic manipulation\cite{xiang2018posecnn}, and navigation\cite{song2015sun}. Inverse graphics is an ``analysis-by-synthesis" approach to 3D scene understanding that has found successful applications in a wide variety of tasks~\cite{hejrati2014analysis,isola2013scene,gall2008drift,gothoskar20213dp3,krull2015learning}. By synthesizing images from possible 3D descriptions of the scene and selecting the 3D scene description that best agrees with the observed image, inverse graphics offers an intuitive and appealing way to reason about the 3D structure of a scene from 2D images. However, challenges such as modeling the gap between rendered images and real-world observations and efficient inference have limited the widespread usage of 3D inverse graphics.

In this paper, we focus on 6D pose estimation, an important task in 3D scene understanding using inverse graphics that aims to infer the rigid $\mathbb{S}\mathbb{E}(3)$ transformations (position and orientation) of objects in the camera frame given an image observation. We emphasize principled probabilistic modeling as a way to address the central challenges in 3D inverse graphics, and propose 3D Neural Embedding Likelihood (3DNEL). Instead of naively rendering RGB images, 3DNEL uses learned neural embeddings to predict 2D-3D correspondences from RGB and combines this with depth to robustly evaluate the agreement of scene descriptions and real-world observations. This results in a unified probabilistic model over RGB-D images that jointly models multiple objects in a scene. We additionally develop efficient inference procedures using 3DNEL, both with stochastic search for 6D object pose estimation from static RGB-D images, and with particle filtering for object pose tracking from video.

We conduct extensive experiments on the popular YCB-Video (YCB-V) dataset~\cite{xiang2018posecnn}. Our results demonstrate that 3DNEL's probabilistic formulation addresses 3D inverse graphics' central challenges of bridging the gap between rendered images and real-world observations, significantly improving robustness in sim-to-real 6D pose estimation on challenging scenes with principled pose uncertainty quantification, while achieving accuracy on par with state-of-the-art (SOTA) approaches that require extensive tuning. Additionally, 3DNEL's joint modeling of multiple objects in a scene and natural support for uncertainty quantification enables robust object pose tracking under occlusion. Furthermore, 3DNEL's probabilistic formulation provides a principled framework for incorporating prior knowledge about the scene and objects, enabling easy extension to additional tasks like camera pose tracking from video, using principled inference in the same probabilistic model without task-specific retraining.

While the field of 6D pose estimation is currently dominated by discriminative approaches based on deep learning, our probabilistic inverse graphics approach provides a complementary alternative that offers unique advantages in terms of robustness, uncertainty quantification and support for multiple tasks due to its probabilistic generative formulation. Our main contributions are three-fold:
\begin{itemize}[nosep]
	\item We propose a probabilistic inverse graphics approach to 6D pose estimation that can naturally support uncertainty quantification, track object poses with particle filtering, and incorporate additional knowledge about the scene and objects to handle camera pose tracking without task-specific retraining.
	\item We conduct extensive experiments on YCB-V and perform on par with SOTA while improving robustness with significantly fewer large-error predictions. \item We show 3DNEL can handle challenging cases such as identifying pose uncertainties for symmetric objects and object pose tracking under heavy occlusion.
\end{itemize}

\section{Related Work}

\textbf{3D inverse graphics}\quad Our method follows a long line of work in the \textit{analysis-by-synthesis} paradigm that treats perception as the inverse problem to computer graphics~\cite{knill1996introduction, yuille2006vision, lee2003hierarchical, kersten2004object, mansinghka2013approximate, kulkarni2015picture,kundu20183d}. While conceptually appealing, robustly modeling the gap between the rendered images and real-world observations, especially using appearance information, remains challenging in 3D inverse graphics. Moreover, without uncertainty estimates, even small errors in 3D scene description estimations can be catastrophic for downstream tasks. In recent years, there has been growing interests in leveraging probabilistic formulations in an inverse graphics approach for shape and scene modeling with principled uncertainty quantification~\cite{gothoskar20213dp3,hoffman2022probnerf}. Our work builds on this trend, and additionally integrates appearance modeling through learned dense 2D-3D correspondences\cite{hodan2020epos,li2019cdpn, he2020pvn3d,florence2018dense, haugaard2022surfemb} with depth information in a unified probabilistic framework to allow superior sim-to-real transfer.

\textbf{Discriminative 6D object pose estimation} Discriminative approaches based on deep learning have recently yielded strong performance on 6D object pose estimation. Existing methods either directly regress poses~\cite{xiang2018posecnn,wang2019densefusion,he2021ffb6d,wang2021gdr}, or first establish 2D-3D correspondences~\cite{hodan2020epos,li2019cdpn,park2019pix2pose,he2020pvn3d,haugaard2022surfemb} followed by a variant of Perspective-n-point (PnP) and random sample consensus (RANSAC)~\cite{fischler1981random}. While such approaches achieve impressive results on a wide variety of datasets, their discriminative nature means there is no natural way to quantify uncertainty, and they cannot be easily extended beyond object pose estimation to additional tasks like object or camera pose tracking from video.

\textbf{Neural embeddings for dense correspondence} Many pose estimation methods directly regress 3D object coordinates at each pixel~\cite{hodan2020epos,li2019cdpn,park2019pix2pose,shugurov2021dpodv2} to predict dense 2D-3D correspondences. Recent works~\cite{florence2018dense,neverova2020continuous} show that we can instead learn neural embeddings for 2D pixel and 3D surface locations, and use the embedding similarities to establish dense correspondence. Several recent pose estimation methods~\cite{haugaard2022surfemb,zhang2022self} demonstrate the benefits of this approach for symmetry handling and category-level generalization. We observe we can combine such embedding similarities with a noise model on the depth information into a unified probabilistic model on RGB-D images. Specifically, we build on SurfEMB~\cite{haugaard2022surfemb}, and show how the additional probabilistic modeling improves both robustness and accuracy while additionally allowing principled uncertainty quantification and easy extension to additional tasks.

\textbf{Render-and-compare for pose refinement} Several recent works~\cite{labbe2020cosypose,labbe2022megapose,li2018deepim,manhardt2018deep,lipson2022coupled} adopt a render-and-compare approach for pose refinement, which resembles the idea of ``analysis-by-synthesis'' in an inverse graphics approach. However, these methods are all discriminative in nature, and train neural networks that take the rendered and real images as inputs, and either directly predict the pose transformations~\cite{labbe2020cosypose,labbe2022megapose,li2018deepim,manhardt2018deep} or predict a flow field~\cite{lipson2022coupled}. In contrast, 3DNEL adopts a probabilistic generative formulation which allows natural support for uncertainty quantification and multiple tasks using principled inference within the same probabilistic model. Moreover, existing render-and-compare methods all consider different objects separately, while 3DNEL jointly models multiple objects in a scene.

\textbf{Sim-to-real transfer} Recent advances in photorealistic rendering and physics-based simulations \cite{hodan2018bop,denninger2019blenderproc} and domain randomization~\cite{tobin2017domain} have yielded impressive results~\cite{haugaard2022surfemb,wang2021gdr,liu2022gdrnpp_bop} in sim-to-real 6D object pose estimation. 3DNEL builds on such advances, and demonstrates that principled probabilistic modeling of the noise distribution between rendered and real-world data can further improve robustness and accuracy in sim-to-real transfer.

\textbf{Uncertainty Quantification and Pose Tracking}\quad Several works \cite{deng2021poserbpf,okorn2020learning,lu2021consensus,sundermeyer2018implicit} propose to quantify pose uncertainties, especially for rotations, to achieve robust performance in ambiguous settings such as symmetric objects and heavily occluded scenes. In our work, we demonstrate how 3DNEL naturally supports such uncertainty quantification, and additionally show how this helps enable the challenging task of object pose tracking under occlusion.


\section{Methods}

\subsection{Preliminaries}\label{sec:preliminaries}



\noindent\textbf{Probablistic inverse graphics}\quad 3D inverse graphics formulates the perception problem as searching for the 3D scene description that can be rendered by a graphics engine to best reconstruct the input image.  We propose a likelihood $\mathbb{P}(\textit{Observed RGB-D Image} | \textit{3D scene description})$ that can assess how well an observed RGB-D image is explained by a 3D scene description.
We define a 3D scene description in terms of the number $N$ of objects in the scene, their classes $t_1, \cdots, t_N \in \{1, \cdots, M\}$, and their corresponding poses $\mathcal{D} = (\mathbf{P}_1, \cdots, \mathbf{P}_N)$ where $\mathbf{P}_1, \cdots, \mathbf{P}_N \in \mathbb{SE}(3)$. Each object is associated with a textured mesh, which captures the 3D shape and appearance information of the object. We assume uniform prior distributions over object poses (uniform over a bounded volume for position and uniform on $\mathbb{SO}(3)$ for orientation). Note that our probabilistic formulation jointly models all objects in the scene, as opposed to many existing probabilistic models where different objects are considered separately.

\noindent\textbf{Noise model on depth information}\quad We use the probabilistic model  $ \mathbb{P}_{\text{depth}}(\mathbf{c} | \mathbf{\tilde{c}}; r)=\frac{\mathbf{1}[||\mathbf{c} - \mathbf{\tilde{c}}||_2 \leq r]}{\frac{4}{3}\pi r^3}$ from 3DP3~\cite{gothoskar20213dp3} as our noise model on depth information. $\mathbb{P}_{\text{depth}}$ is a uniform distribution in a radius-$r$ ball centered at a rendered point $\mathbf{\tilde{c}}\in\mathbb{R}^3$, and models the small spatial displacements in the observed point $\mathbf{c}\in\mathbb{R}^3$. $r$ is a hyperparameter that controls the variance of the noise model.

\noindent\textbf{Noise model on RGB information}\quad  Instead of directly operating on RGB images, we leverage similarity measurements of learned neural embeddings for 2D pixel and 3D surface locations~\cite{florence2018dense,neverova2020continuous,haugaard2022surfemb,zhang2022self} to specify the noise model on RGB information. As a concrete example, we reuse components from SurfEMB~\cite{haugaard2022surfemb} to highlight how 3DNEL's principled probabilistic modeling brings added benefits on robustness and uncertainty quantification. But in Section~\ref{sec:dino} we illustrate how 3DNEL can leverage other learned neural embeddings and similarity measurements.

For each object class $t\in \{1, \cdots, M\}$, SurfEMB learns two neural embedding models: (1) a \textit{query embedding model} which maps an RGB image $\mathbf{I}$ to a set of query embedding maps $\mathbf{Q}^t$, one for each object class, and (2) a \textit{key embedding model} $g_t: \mathbb{R}^{3} \mapsto \mathbb{R}^{E}$ which maps each 3D location $\mathbf{x}\in\mathbb{R}^3$ (object frame coordinate) on the object surface to a key embedding $g_t(\mathbf{x})\in\mathbb{R}^E$. Given a pixel with query embedding $\mathbf{q} \in \mathbb{R}^{E}$, SurfEMB measures the similarity between the query and the key embeddings using a surface distribution $\mathbb{P}_{\text{RGB}}(g_t(\mathbf{x}) |\mathbf{q}, t) \propto \exp(\mathbf{q}^T g_t(\mathbf{x}))$ that describes which point $\mathbf{x}$ on the object surface the given
pixel corresponds to. Importantly, these models can be trained entirely from synthetic data (with photorealistic rendering, physics-based simulations and domain randomization). See Appendix A for a more detailed review.




\subsection{3D Neural Embedding Likelihood (3DNEL)}\label{sec:nel}

\begin{figure*}[t!]
	\centering
	\includegraphics[width=\textwidth]{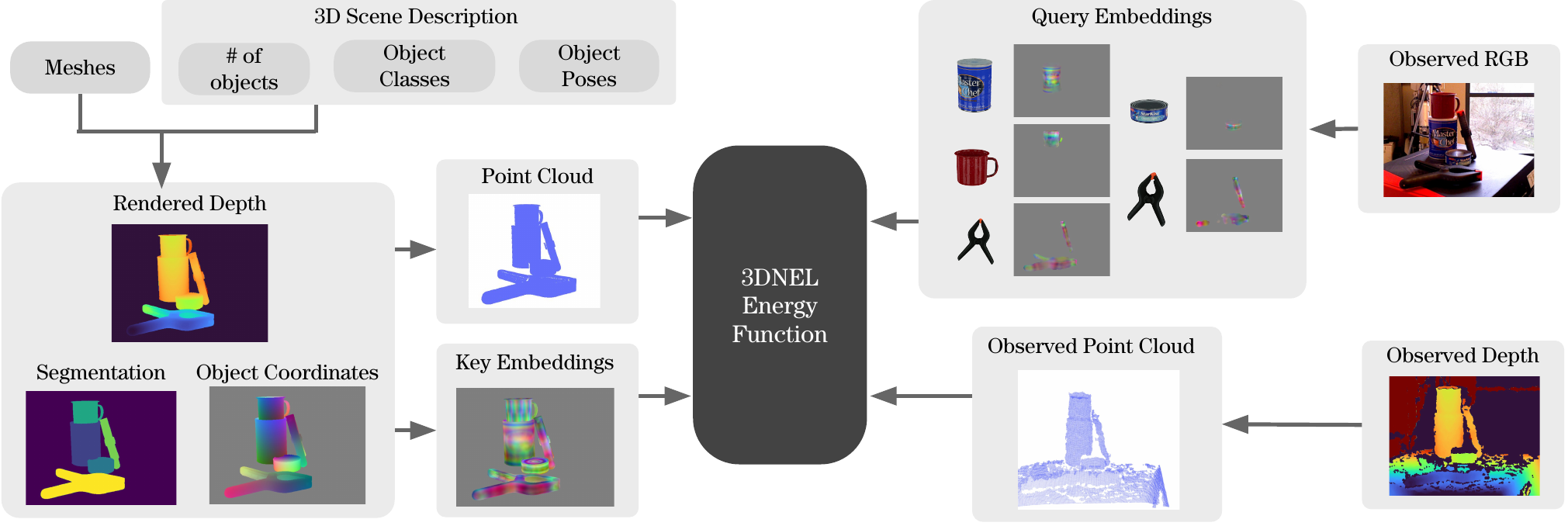}
	\caption{\textbf{Evaluating 3DNEL} 3DNEL defines the probability of an observed RGB-D image conditioned on a 3D scene description. We first render the 3D scene description into: (1) a depth image, which is transformed to a rendered point cloud image, (2) a semantic segmentation map, and (3) the object coordinate image (each pixel contains the object frame coordinate of the object surface point from which the pixel originates). The object coordinate image is transformed, via the key models, into key embeddings. The observed RGB image is transformed, via the query models, into query embeddings. The observed depth is transformed into an observed point cloud image. The 3DNEL Energy Function (Equation~\ref{eqn:energy_based}) is evaluated using the rendered point cloud image, semantic segmentation, key embeddings, the observed point cloud image, and query embeddings.}
	\label{fig:likelihood}
\end{figure*}

\noindent\textbf{Processing 3D scene description for 3DNEL evaluation}\quad For a given 3D scene description, we use a 3D graphics engine to render it into: (1) A rendered point cloud image $\mathbf{\tilde{C}}$, where $\mathbf{\tilde{C}}_{i, j} \in \mathbb{R}^3$ represents the camera frame coordinate at pixel $(i, j)$. (2) A semantic segmentation map $\mathbf{\tilde{S}}$ where $\mathbf{\tilde{S}}_{i, j}\in\{0, 1, \cdots, M\}$ represents the class to which the pixel $(i, j)$ belongs. Here $0$ represents background. (3) An object coordinate image $\mathbf{\tilde{X}} $ where $\mathbf{\tilde{X}}_{i, j}\in\mathbb{R}^3$ represents the object frame coordinate at pixel $(i, j)$ of the object of class $\mathbf{\tilde{S}}_{i, j}$.

\noindent\textbf{Processing RGB-D image for 3DNEL evaluation}\quad For an observed RGB image $\mathbf{I}$ and depth image, we use the learned query embedding models to obtain $M$ sets of query embedding maps $\mathbf{Q}^t, t\in\{1, \cdots, M\}$, one for each object class, where $\mathbf{Q}^t_{i,j}\in\mathbb{R}^E$ represents the query embedding at pixel $(i,j)$, and use camera intrinsics to unproject the depth image into an observed point cloud image $\mathbf{C}$, where $\mathbf{C}_{i,j}\in\mathbb{R}^3$ represents the camera frame coordinate at pixel $(i,j)$.

\noindent\textbf{3DNEL evaluation}\quad Figure~\ref{fig:likelihood} visualizes 3DNEL evaluation using processed 3D scene descriptions and observed RGB-D images. 3DNEL combines the noise model $\mathbb{P}_{\text{depth}}$ on depth information and the dense 2D-3D correspondence distribution $\mathbb{P}_{\text{RGB}}$, and jointly models multiple objects in a scene through a mixture model formulation. This results in a unified probabilistic model on real RGB-D images.

Intuitively, we assess how well each pixel $(i, j)$ in the observed point cloud image $\mathbf{C}$ is explained by a pixel $(\tilde{i}, \tilde{j})$ in the rendered point cloud $\mathbf{\tilde{C}}$, by combining the noise model $\mathbb{P}_{\text{depth}}$ on depth and the noise model $\mathbb{P}_{\text{RGB}}$ on RGB. To jointly model multiple objects in a scene, we assume each pixel $(i, j)$ in $\mathbf{C}$ can be explained by multiple pixels in $\mathbf{\tilde{C}}$.

We formalize this with a mixture model formulation, where the mixture component associated with the rendered pixel $(\tilde{i}, \tilde{j})$ combines $\mathbb{P}_{\text{depth}}$ and $\mathbb{P}_{\text{RGB}}$ to assess how well the observed pixel $(i, j)$ is explained by the rendered pixel $(\tilde{i}, \tilde{j})$. To model background pixels in $\mathbf{C}$, we assume the observed point cloud image $\mathbf{C}$ resides in a bounded region of volume $B$, and introduce a uniform distribution $\mathbb{P}_{\text{BG}}(\mathbf{c}; B) = 1/B$ on the bounded region with mixture probability $\epsilon$ as an additional mixture component for background modeling. Representing the total number of non-background pixels in the rendered images as $\tilde{K} = \sum_{\tilde{i}, \tilde{j}}\mathbf{1}[\mathbf{\tilde{S}}_{\tilde{i}, \tilde{j}} > 0]$, the mixture probability for the mixture component associated with rendered pixel $(\tilde{i}, \tilde{j})$ is given by $(1 - \epsilon) / \tilde{K}$. Since the query embedding at a pixel depends on the entire image $\mathbf{I}$, the mixture components are not properly normalized. This leads to the following energy-based formulation: $\mathbb{P}_{\text{3DNEL}}(\mathbf{I}, \mathbf{C} | \mathcal{D})$ is proportional to
\begin{equation}\label{eqn:energy_based}\footnotesize
	\prod_{\mathbf{c}}
	\left( \epsilon\mathbb{P}_{\text{BG}}(\mathbf{c}; B) + \frac{1 - \epsilon}{\tilde{K}}\sum_{\mathbf{\tilde{c}}:\tilde{s} > 0}
	\mathbb{P}_{\text{depth}}(\mathbf{c} | \mathbf{\tilde{c}}; r)
	\mathbb{P}_{\text{RGB}}(g_{\tilde{s}}(\mathbf{\tilde{x}}) | \mathbf{q}^{\tilde{s}}, \tilde{s}) \right)
\end{equation}
where we denote $\mathbf{C}_{i, j}$ by $\mathbf{c}$, $\mathbf{\tilde{C}}_{\tilde{i}, \tilde{j}}$ by $\mathbf{\tilde{c}}$, $\mathbf{\tilde{S}}_{\tilde{i}, \tilde{j}}$ by $\tilde{s}$, $\mathbf{\tilde{X}}_{\tilde{i}, \tilde{j}}$ by $\tilde{x}$, and $\mathbf{Q}^t_{i, j}$ by $\mathbf{q}^t$. The product is over all observed pixels, and the sum is over all non-background rendered pixels. $\epsilon, B$ and $r$ are hyper-parameters that we pick in the experiments. See Appendix B for more details.

\subsection{Inferring the 3D scene description}\label{sec:inference}

\begin{figure*}[t]
	\centering
	\includegraphics[width=0.9\textwidth]{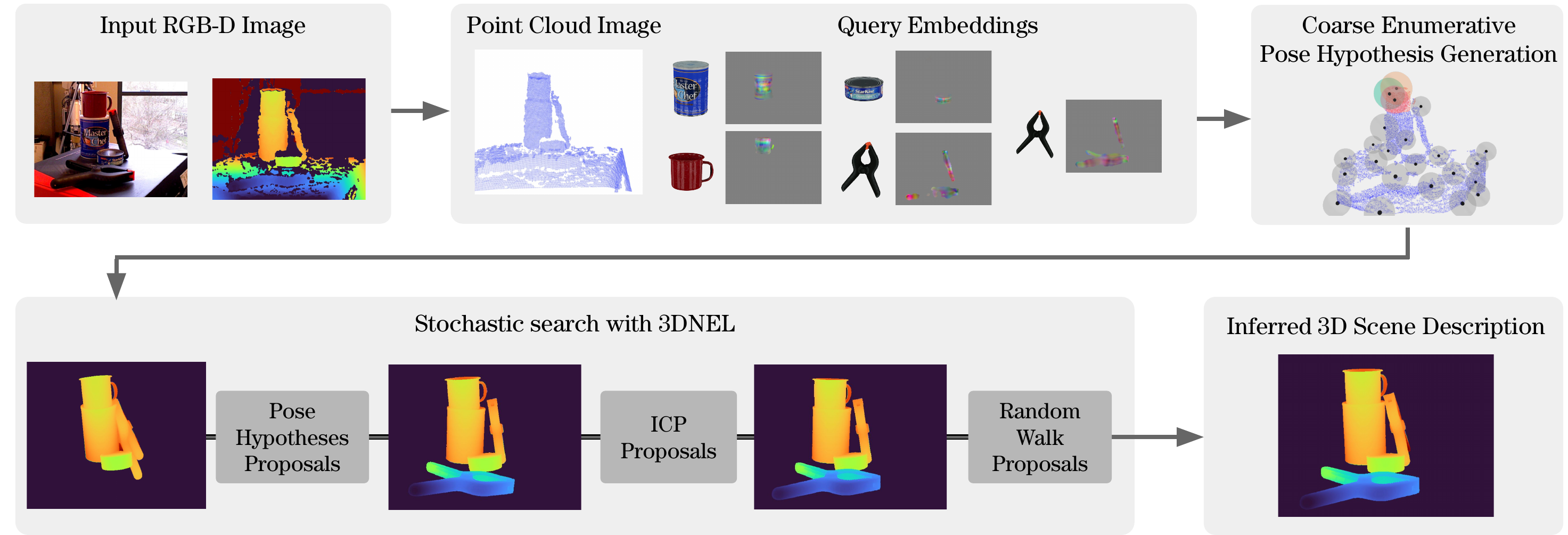}
	\caption{\textbf{Using 3DNEL for 3D Scene Parsing} The 3DNEL MSIGP pipeline starts by computing the query embeddings for each object and the observed point cloud image from RGB-D observations. Then, a fast enumerative procedure produces the pose hypotheses for the objects, and construct an initial 3D scene description. We further perform stochastic search with 3DNEL using three types of MH proposals (1) pose hypotheses proposals  (2) ICP proposals to align an object to point cloud data, and (3) random walk proposals that refines poses with local perturbations. The result is a 3D scene description that explains the observed RGB-D image. 3DNEL's joint modeling of multiple objects through the mixture model formulation enables robust estimation on this challenging scene with two similar-looking clamps.}
	\label{fig:pipeline}
\end{figure*}

\noindent\textbf{Stochastic search with 3DNEL}\quad Given an observed RGB-D image (represented as $\mathbf{I}$ for the RGB image and $\mathbf{C}$ for the observed point cloud) and a 3D scene description (with object poses $\mathcal{D} = (\mathbf{P}_1, \cdots, \mathbf{P}_N)$), 3DNEL evaluates the likelihood $\mathbb{P}(\mathbf{I}, \mathbf{C} | \mathcal{D})$ using Equation~\ref{eqn:energy_based} as described in Section~\ref{sec:nel}. We develop an OpenGL-based parallel renderer, and a JAX~\cite{jax2018github} based likelihood evaluation using the rendered outputs. This allows efficient parallel evaluation of the likelihood of an observed RGB-D image for hundreds of 3D scene descriptions on modern GPUs.

We design a stochastic search procedure with 3DNEL to infer the 3D scene description from an observed RGB-D image. Given the current 3D scene description $\mathcal{D}$, the stochastic search proceudre is an iterative process where at each iteration, we propose $K$ candidate poses $\mathbf{\tilde{P}}_1, \cdots, \mathbf{\tilde{P}}_K$ for a randomly picked object $i\in\{1, \cdots, N\}$. We evaluate in parallel the likelihood of $K$ 3D scene descriptions obtained by replacing the pose $\mathbf{P}_i$ of object $i$ in $\mathcal{D}$ with each of the $K$ candidate poses, and identify the candidate pose with the highest likelihood. We update $\mathbf{P}_i$ to this candidate pose if this increases the likelihood.

We consider 3 types of pose proposals: (1) pose hypotheses proposal proposes a pre-specified set of pose hypotheses (obtained either from the coarse enumerative procedure in Section~\ref{sec:msigp} or from a different pose estimation method); (2) ICP proposal uses ICP to align the object to the observed point cloud, and proposes a set of candidate poses sampled from a Gaussian-von Mises–Fisher (Gaussian-VMF) distribution centered around the aligned pose; (3) random walk proposal proposes a set of candidate poses sampled from a Gaussian-VMF distribution centered around the object's current pose. The Gaussian-VMF distribution means a multivariate Gaussian centered at the current position and a VMF distribution centered at the current orientation.

Given a set of coarse pose estimations as pose hypotheses, the stochastic search procedure with 3DNEL can be used for pose refinement. As we demonstrate in Section~\ref{sec:sim-to-real}, 3DNEL's joint modeling of multiple objects in a scene and principled combination of RGB and depth information allows such pose refinement process to further improve robustness and accuracy of previous SOTA.

\noindent\textbf{Particle filtering for object pose tracking from video}\quad We formulate the problem of object pose tracking from video as probabilistic inference in a state-space model. At each timestep $t=1,\dots,T$, we have the 3D scene description $\mathcal{D}_t = (\mathbf{P}^{(t)}_1, \cdots, \mathbf{P}^{(t)}_N)$ as the latent state and the RGB-D image $\mathbf{I}_t, \mathbf{C}_t$ as the observed variable. We use a simple dynamics model $\mathbb{P}_{\text{dynamics}}(\mathcal{D}_{t + 1} | \mathcal{D}_t)$ that independently samples the poses of each object at time $t+1$ from Gaussian-VMF distributions centered at the poses of the objecs at time $t$, and use 3DNEL as the likelihood. We have the following state space model $\mathbb{P}(\mathcal{D}_{1:T}, \mathbf{I}_{1:T}, \mathbf{C}_{1:T})$:
\begin{equation}
	\mathbb{P}(\mathcal{D}_1) \prod_{t=1}^{T - 1} \mathbb{P}_{\text{dynamics}}(\mathcal{D}_{t+1} | \mathcal{D}_{t}) \prod_{t=1}^T \mathbb{P}_{\text{3DNEL}}(\mathbf{I}_t, \mathbf{C}_t | \mathcal{D}_t)
\end{equation}
Given a sequence of RGB-D frames from a video, we use the Sampling Importance Resampling (SIR) particle filter~\cite{gordon1993novel,arulampalam2002tutorial} to infer the posterior distribution $\mathbb{P}(\mathcal{D}_{1:T} | \mathbf{I}_{1:T}, \mathbf{C}_{1:T})$ , and use $\argmax_{\mathcal{D}_t} \mathbb{P}(\mathcal{D}_t | \mathbf{I}_{1:t}, \mathbf{C}_{1:t})$ as our tracking estimate at time $t$.

\subsection{6D object pose estimation pipeline}\label{sec:msigp}
\noindent\textbf{Coarse Enumerative Pose Hypotheses Generation}\quad Existing pose estimation methods based on dense 2D-3D correspondences typically use PnP to generate pose hypotheses. However, PnP does not take depth information into account, requires the use of the time-consuming RANSAC to deal with noisy 2D-3D correspondences, and needs separate 2D detections to localize and mask out the object.

Motivated by the above, we develop novel spherical voting and heuristic scoring procedures, and use a coarse enumerative procedure to efficiently generate pose hypotheses. Given a set of keypoints sampled from the object surface using farthest point sampling, spherical voting leverages dense 2D-3D correspondences to estimate the 3D distance between an observed point and possibly present keypoints around it, and cast votes towards all points on spheres with the predicted distances as radiuses to aggregate information from the entire RGB-D image into a 3D accumulator space. We coarsely discretize the object pose space, and heuristically score the discretized poses using the aggregated information to output top scoring pose hypotheses. Refer to Appendix C for a detailed description of the process.

Our coarse enumerative pose hypotheses generation combines depth information with dense 2D-3D correspondences, and can be implemented efficiently on the GPU (we use Taichi~\cite{hu2019taichi}). As we show in Section~\ref{sec:sim-to-real}, it performs competitively even without separate 2D detections, and can additionally leverage available 2D detections to filter out noisy query embeddings and restrict voting to only the relevant image regions to further improve performance.

\begin{table}[t]
	\centering
	\renewcommand{\arraystretch}{1.1}
	\resizebox{0.5\textwidth}{!}{
		\begin{tabular}{l|l|c}
			\textbf{Category}                               & \textbf{Method}                                                  & \textbf{Average Recall} \\ \hline
			\multirow{2}{*}{\textbf{Core comparison}}       & 3DNEL MSIGP (Ours)                                               & 84.85\%                 \\
			                                                & SurfEMB \cite{haugaard2022surfemb}                               & 80.00\%                 \\
			\hline
			\multirow{5}{*}{\shortstack[l]{\textbf{Additional baselines}                                                                                 \\\textbf{in the sim-to-real setup}}} & CosyPose \cite{labbe2020cosypose}                                                           & 71.42\%                 \\
			                                                & Coupled Iterative Refinement \cite{lipson2022coupled}            & 76.58\%                 \\
			                                                & FFB6D \cite{he2021ffb6d}                                         & 75.80\%                 \\
			                                                & MegaPose  \cite{labbe2022megapose} (also supports novel objects) & 63.3\%                  \\
			                                                & GDRNPP\cite{liu2022gdrnpp_bop} (concurrent work)                 & 90.6\%                  \\

			\hline
			\multirow{7}{*}{\textbf{3DNEL MSIGP Ablations}} & No RGB in Likelihood                                             & 61.57\%                 \\
			                                                & No Depth in Likelihood                                           & 50.85\%                 \\
			                                                & SurfEMB initialization + stochastic search                       & 82.73\%                 \\
			                                                & No 2D Detection                                                  & 72.08\%                 \\
			                                                & No pose hypotheses proposal                                      & 80.57\%                 \\
			                                                & No ICP proposal                                                  & 81.86\%                 \\
			                                                & No random walk proposal                                          & 78.28\%                 \\
			\hline
		\end{tabular}}

	\caption{\textbf{3DNEL MSIGP achieves accuracy on par with SOTA, and outperforms ablations on YCB-V} We report Average Recall on YCB-V in the sim-to-real setup using RGB-D inputs. Results for 3DNEL MSIGP are averaged over 5 runs. Standard deviation is below $0.2\%$ for all setups. 3DNEL MSIGP significantly outperforms SurfEMB depiste using the same underlying models, highlighting the benefits of 3DNEL's principled probabilistic modeling. In addition, 3DNEL MSIGP achieves results that are on par with SOTA, and outperforms all the included baselines in the sim-to-real setup by a large margin, except concurrent work GDRNPP~\cite{liu2022gdrnpp_bop} which achieves a new SOTA but requires extensive tuning in terms of 2D detection, backbone architectures, data augmentation, training hyper-parameters and pose refinement~\cite{lipson2022coupled}.}
	\label{tab:average_recall}
	\vspace*{-10pt}
\end{table}

\noindent\textbf{3DNEL multi-stage inverse graphics pipeline (MSIGP)}\quad We design a MSIGP based on 3DNEL for sim-to-real 6D object pose estimation. We generate a set of pose hypotheses for each object class using the above coarse enumerative procedure, and initialize the 3D scene description with the top scoring pose hypothesis for each object class. Starting from the initial 3D scene description, we use the stochastic search procedure as described in Section~\ref{sec:inference} to infer the 3D scene description
\begin{equation*}
	\mathbf{\tilde{P}}_1, \cdots, \mathbf{\tilde{P}}_N = \argmax_{\mathbf{P}_1, \cdots, \mathbf{P}_N} \mathbb{P}(\mathbf{I}, \mathbf{C}| \mathbf{P}_1, \cdots, \mathbf{P}_N)
\end{equation*}
We start with the pose hypotheses proposal, followed by the ICP proposal, before finally applying the random walk proposal. For each type of proposal, we go through all the objects once. See Figure~\ref{fig:pipeline} for an illustration of the 3DNEL MSIGP. We follow \cite{he2021ffb6d} and use \cite{ku2018defense} to fill in missing depth. See Appendix D for details.

\noindent\textbf{Training}\quad To highlight that performance improvements are coming entirely from 3DNEL's probabilistic formulation, we use publicly released pretrained SurfEMB models, and also use SurfEMB as our primary baseline. These include query models, key models, mask predictors for different object classes, and an additional 2D detector from CosyPose~\cite{labbe2020cosypose}, all trained entirely from synthetic data.

\noindent\textbf{Comparison with SurfEMB}\quad SurfEMB proposes a pose estimation pipeline based on probabilistic modeling of dense 2D-3D correspondences: it samples 2D-3D correspondences on image crops from object detection, generates pose hypotheses using PnP, scores the pose hypotheses using its training objective, and does gradient-based refinement. It does all of the above separately for individual objects using only RGB information. Depth is only used in median depth comparison for heuristic pose refinement.

In contrast, 3DNEL's probabilistic inverse graphics approach focuses on explaining the entire scene, and the proposed 3DNEL MSIGP differs from SurfEMB's pose estimation pipeline in a few key ways: (1) 3DNEL MSIGP jointly models all objects which improves robustness. (2) 3DNEL MSIGP generates pose hypotheses using a coarse enumerative procedure, which we demonstrate to be faster and better in Section~\ref{sec:exps}. (3) 3DNEL MSIGP combines correspondence and depth in a principled way in all aspects of the modeling (likelihood formulation, hypotheses generation and refinement) and improves performance.

\section{Experiments}\label{sec:exps}

\begin{figure*}[t!]
	\centering
	\includegraphics[width=\linewidth]{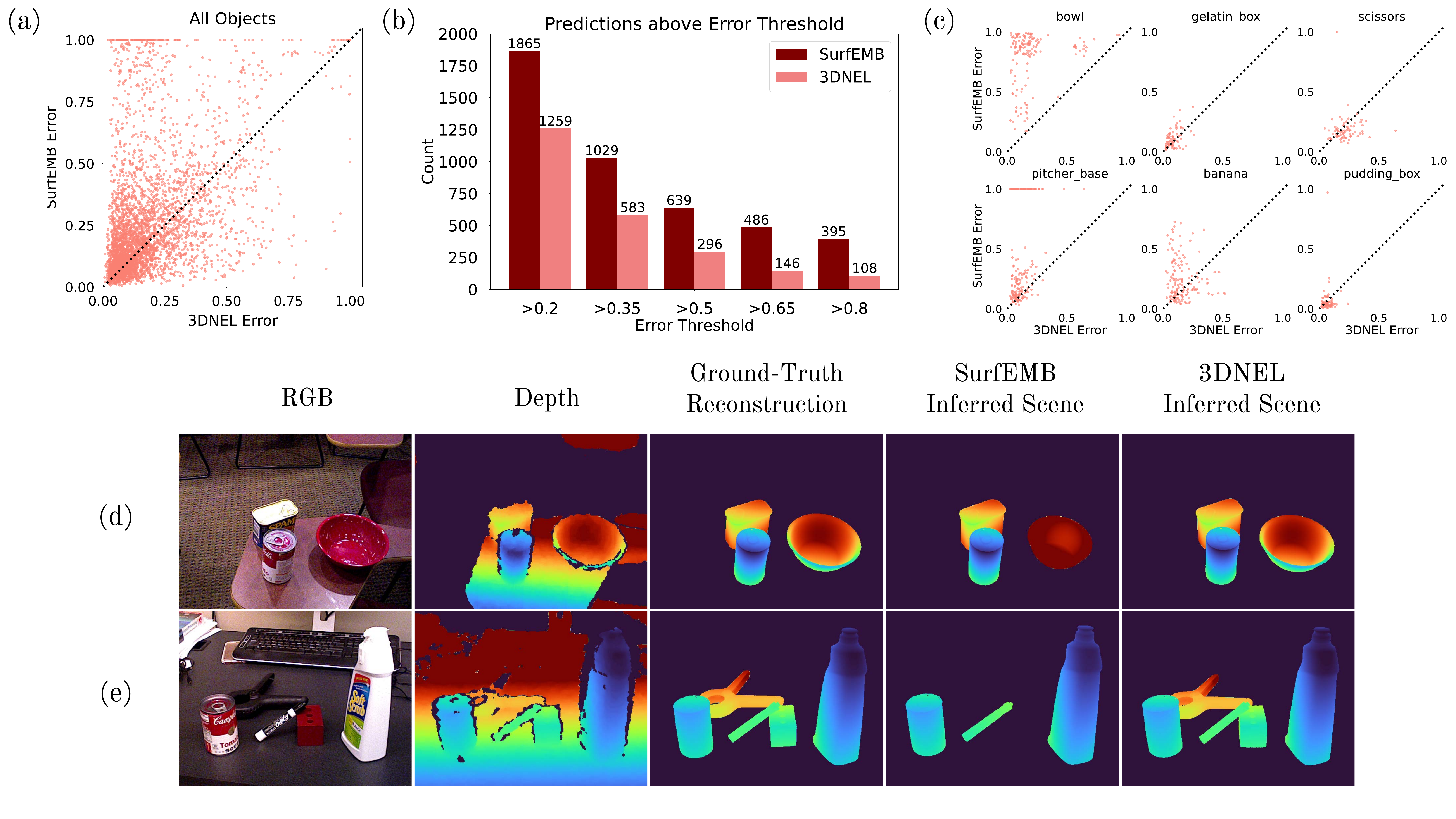}
	\caption{\textbf{3DNEL MSIGP improves robustness over SurfEMB} (a) Comparison of prediction error (measured by VSD) between SurfEMB and 3DNEL MSIGP across 4123 object instances in YCB-V. Each point on the scatter plot represents an instance. Points above the dashed line represent instances for which 3DNEL MSIGP has lower prediction error. (b) Number of instances with prediction error above a certain error threshold, across multiple thresholds. 3DNEL MSIGP makes significantly less high-error predictions than SurfEMB (over 50\% less above 0.5). (c) Scatter plots for 6 representative object classes. (d)(e) 3DNEL MSIGP is more robust than SurfEMB on challenging scenes.}
	\label{fig:error_scatter_plots}
	\vspace*{-10pt}
\end{figure*}

In this section, we aim to answer the following questions: (1) Can 3DNEL achieve improved robustness in challenging sim-to-real setups compared to more discriminative baselines? (2) Does 3DNEL perform on par with SOTA? (3) Can 3DNEL be additionally used to quantify uncertainty, as well as for object and camera pose tracking?

\subsection{Sim-to-real object pose estimation\footnote{Code to reproduce the results in this section is available at \url{https://github.com/deepmind/threednel}.}}\label{sec:sim-to-real}



\noindent\textbf{Evaluation}\quad We follow the evaluation protocol of the Benchmark for 6D Object Pose Estimation (BOP) challenge \cite{hodan2018bop}. The task is to estimate the 6D poses of objects in a scene from a single RGB-D image, assuming knowledge of the number of instances of each object class in the scene. For a predicted pose, we calculate three error metrics: Visible Surface Discrepancy (VSD) \cite{hodavn2016evaluation,hodan2018bop},  Maximum Symmetry-Aware Surface Distance (MSSD) \cite{drost2017introducing}, and Maximum Symmetry-Aware Projection Distance \cite{brachmann2016uncertainty}. Average recalls $\text{AR}_{VSD}$, $\text{AR}_{MSSD}$, $\text{AR}_{MSPD}$ are computed for each error metric across a range of error thresholds. The aggregate Average Recall (as reported in Table \ref{tab:average_recall}) is the average of $\text{AR}_{VSD}$, $\text{AR}_{MSSD}$, and $\text{AR}_{MSPD}$.

\noindent\textbf{Baselines}\quad We use SurfEMB as our main baseline for a series of detailed analysis. We additionally include the sim-to-real performance of several recent SOTA 6D object pose estimation method~\cite{labbe2020cosypose,lipson2022coupled,he2021ffb6d,labbe2022megapose,liu2022gdrnpp_bop} as context. For \cite{labbe2020cosypose,lipson2022coupled}, we use the publicly available codebase to retrain on only synthetic data, and re-evaluate their performance.

\noindent\textbf{Robustness}\quad Figure \ref{fig:error_scatter_plots} illustrates how 3DNEL's probabilistic formulation significantly reduces large-error pose estimations when compared with SurfEMB and improves robustness. Across all YCB-V test images, there are 4123 object instances. Each point on the scatter plots corresponds to an object instance, and the point's $x$ and $y$ coordinates are the pose prediction error of 3DNEL MSIGP and SurfEMB, respectively. Points above the dashed line correspond to object instances for which 3DNEL MSIGP had a lower prediction error than SurfEMB. Figure \ref{fig:error_scatter_plots}(a) shows the scatter plot for all 4123 predictions and Figure \ref{fig:error_scatter_plots}(c) shows the scatter plots for 6 representative object classes. Figure \ref{fig:error_scatter_plots}(b) shows, across a range of error thresholds, the number of pose predictions with error above that threshold.

Qualitatively, we observe 3DNEL's probabilistic formulation is especially helpful in challenging situations: (1) Scenes (e.g. Figure~\ref{fig:pipeline}) with similar-looking objects. SurfEMB makes per-object predictions and incorrectly predicts both clamps to be in the back, while 3DNEL jointly models all objects in the scene and makes correct prediction. (2) Scenes (e.g. Figure~\ref{fig:error_scatter_plots}(d)) with objects like the red bow where RGB alone is not informative enough. With a principled combination of RGB and depth information, 3DNEL MSIGP can reliably correct a large number of related errors. (3) Scenes (e.g. Figure~\ref{fig:error_scatter_plots}(e)) with missing 2D detections. SurfEMB cannot recover from missing 2D detections, while 3DNEL MSIGP can robustly aggregate information from the entire image and reliably avoid such errors.

\begin{figure*}[t!]
	\centering
	\includegraphics[width=0.9\textwidth]{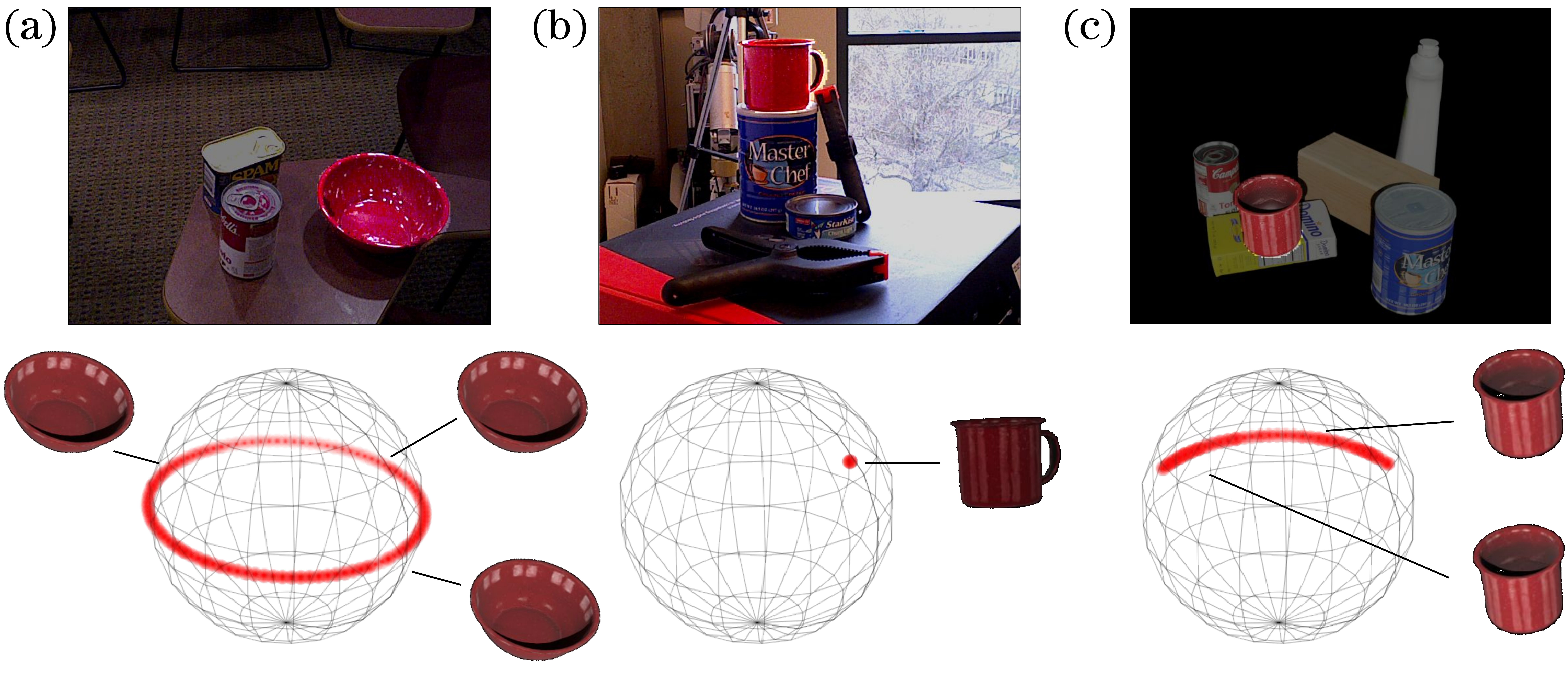}
	\caption{\textbf{3DNEL naturally quantifies pose uncertainty in the scene.} 3DNEL identifies pose uncertainty for the red bowl due to its inherent symmetry, and accurately captures the range of equally likely poses for the red mug when its handle is not visible.}
	\label{fig:pose_uncertainty}
	\vspace*{-10pt}
\end{figure*}

\noindent\textbf{Identifying pose uncertainties}\quad {Pose uncertainty} may arise from partial observability, viewpoint ambiguity, and inherent symmetries of the object. 3DNEL can naturally quantify such pose uncertainties due to its probabilistic formulation. Figure~\ref{fig:pose_uncertainty}(a) illustrates how 3DNEL identifies the pose uncertainty of the red bowl due to its inherent symmetry. Figures~\ref{fig:pose_uncertainty}(b)(c) consider the red mug in YCB objects. 3DNEL can accurately capture that while there is no pose uncertainty when the mug handle is visible, there are a range of equally likely poses when the mug handle is not visible.

\noindent\textbf{Comparison with baselines}\quad In Table \ref{tab:average_recall}, we report the Average Recall for 3DNEL MSIGP and representative recent baselines. 3DNEL MSIGP significantly outperforms SurfEMB depiste using the same underlying models, highlighting the benefits of 3DNEL's principled probabilistic modeling. In addition, 3DNEL MSIGP achieves results that are on par with SOTA, and outperforms all the included baselines in the sim-to-real setup by a large margin, except concurrent work GDRNPP~\cite{liu2022gdrnpp_bop} which requires extensive tuning.

\noindent\textbf{Abalations}\quad We report additional abalations in Table \ref{tab:average_recall}.

We evaluate \textit{No RGB in Likelihood}, where we drop $\mathbb{P}_{\text{RGB}}$ in Equation \ref{eqn:energy_based}, and \textit{No Depth in Likelihood} where we replace $\mathbf{1}[||\mathbf{C}_{i,j} - \mathbf{\tilde{C}}_{\tilde{i}, \tilde{j}}||_2 \leq r]$ in Equation \ref{eqn:energy_based} with a fixed $3\times 3$ 2D patch. Both are substantially worse.

We evaluate \textit{SurfEMB initialization + stochastic search}, which replaces the coarse enumerative pose hypotheses generation with pose hypotheses from SurfEMB, and initializes with SurfEMB predictions. This leads to $2.73\%$ improvement over SurfEMB but is still worse than 3DNEL MSIGP, illustrating the value of our stochastic search and coarse enumerative pose hypotheses generation.

We evaluate \textit{No 2D Detection}, where we do not use 2D detections for pose hypotheses generation. This uses much less information than all the other baselines, yet outperforms strong baselines like CosyPose in the sim-to-real setup, demonstrating the ability of our coarse enumerative pose hypotheses generation to robustly aggregate information from entire noisy query embedding images.

We evaluate the effects of the three different proposals used in our stochastic search, by separately removing the pose hypotheses proposal, ICP proposal and random walk proposal (\textit{No pose hypotheses/ICP/random walk proposal}). Performances remain strong, but drop to various extent, illustrating the contributions from all three proposals.

\noindent\textbf{Inference speed}\quad Inverse graphics approaches are traditionally computationally expensive. To speed up inference, we downsample input images by a factor of 0.25. We fully leverage recent hardware advances to develop efficient GPU implementations. We use a single NVIDIA A100 GPU for our experiments. When tested in the same setup, SurfEMB reported taking 1.2s for pose hypotheses generation using PnP+RANSAC, while our coarse enumerative pose hypotheses generation takes \textbf{0.2s} with better results. Our stochastic search with 3DNEL runs at a similar speed to SurfEMB's pose refinement, and averages around 1s per object. Most of our implementation (except parallel rendering with OpenGL) is written in Python, using a mix of JAX, PyTorch and Taichi. We empirically observe that inter-communication between different packages creates additional overhead. We expect a compiled implementation using a single framework to further speed up inference.

\noindent\textbf{Inference hyperparameters}\quad For the above experiments, we pick inference hyperparameters\footnote{See \url{https://github.com/deepmind/threednel/blob/main/threednel/bop/detector.py} for a complete list.} by visually inspecting inference results on a small number of real training images outside the test set. To verify 3DNEL can robustly apply to a variety of object types and scene configurations, we evaluate on two additional datasets using the same hyperparameters. On TUD-L~\cite{hodavn2020bop}, 3DNEL improves SurEMB by \textbf{2.7}\% (88.1\% vs 85.4\%). On LM-O~\cite{brachmann2014learning}, due to the small objects in the scenes, we increase the downsampling factor from 0.25 to 0.6. Keeping all other hyperparameters the same, 3DNEL improves SurfEMB by \textbf{0.7}\% (76.7\% vs 76.0\%).

\subsection{Object pose tracking under occlusion}

\begin{figure*}[t!]
	\centering
	\includegraphics[width=\textwidth]{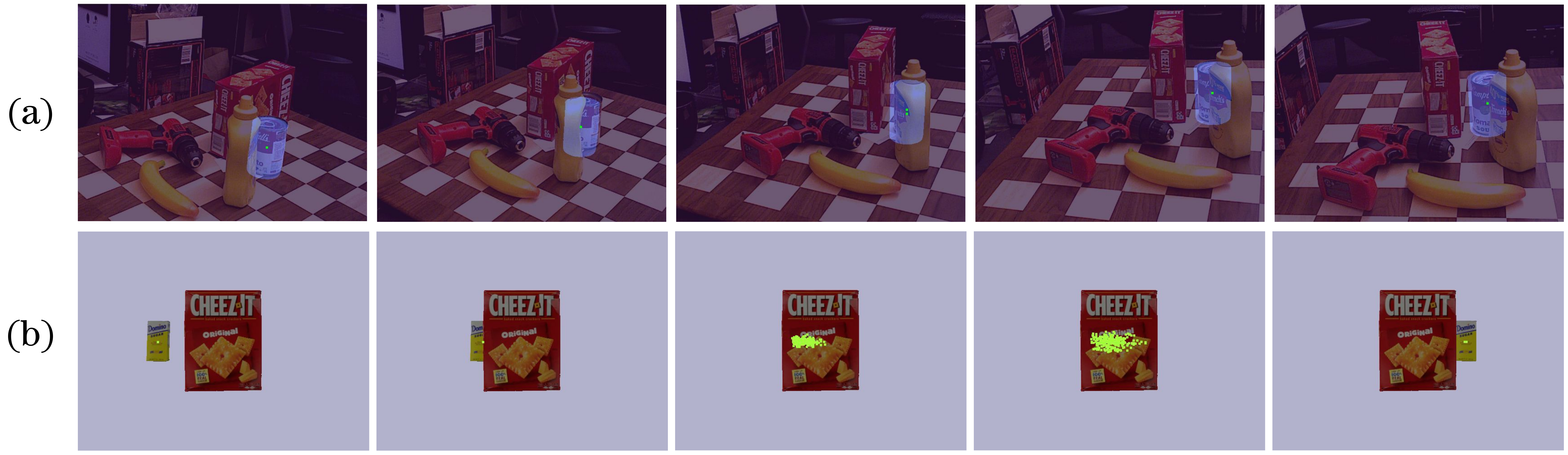}
	\caption{\textbf{3DNEL's probabilistic formulation enables robust object pose tracking under occlusion with particle filtering.} Green dots visualize the particles used to estimate posterior distributions in particle filtering. (a) 3DNEL can track objects through heavy occlusions. (b) 3DNEL can accurately quantify uncertainty (shown by the spread-out particles), which enables tracking through extended occlusions.}
	\label{fig:tracking}
\end{figure*}

\begin{table*}[t!]
	\centering
	\renewcommand{\arraystretch}{1.2}
	\resizebox{1.9\columnwidth}{!}{
		\begin{tabular}{l|cccccccccccc}
			                           & \multicolumn{12}{c}{\textbf{Scene ID}}                                                                                                                                                                                                       \\
			                           & 48                                     & 49              & 50              & 51              & 52              & 53              & 54              & 55              & 56              & 57              & 58              & 59              \\ \hline
			SurfEMB Single Frame       & 77.6\%                                 & 67.0\%          & 83.7\%          & 91.3\%          & 80.0\%          & 59.8\%          & 88.4\%          & 76.7\%          & 70.5\%          & 77.3\%          & 92.4\%          & 84.1\%          \\
			3DNEL MSIGP Single Frame   & 71.9\%                                 & 77.5\%          & 83.1\%          & 87.7\%          & 87.5\%          & 84.1\%          & 88.4\%          & 80.4\%          & 82.8\%          & 85.3\%          & 94.3\%          & 86.4\%          \\
			3DNEL camera pose Tracking & \textbf{81.5\%}                        & \textbf{94.7\%} & \textbf{97.5\%} & \textbf{97.0\%} & \textbf{97.0\%} & \textbf{97.0\%} & \textbf{97.2\%} & \textbf{97.5\%} & \textbf{96.8\%} & \textbf{92.2\%} & \textbf{98.0\%} & \textbf{97.0\%}
		\end{tabular}
	}
	\caption{ \textbf{Extending 3DNEL to camera pose tracking improves performance compared to single-frame setups} We apply 3DNEL to camera pose tracking from video. We demonstrate that 3DNEL's probabilistic formulation allows us to incorporate additional knowledge that the scene is static and leverage temporal information to significantly improve pose estimation accuracy over the single frame setting.}
	\label{tab:tracking}
	\vspace*{-10pt}
\end{table*}

We apply 3DNEL under the particle filtering framework as described in Section~\ref{sec:inference} for object pose tracking under occlusion. Figure~\ref{fig:tracking}(a) visualizes tracking with 3DNEL with 200 particles on a representative YCB-V video, where the tomato can gets fully occluded before reappearing. Existing per-object likelihood~\cite{deng2021poserbpf} cannot handle such cases without ad hoc occlusion modeling. In contrast, 3DNEL's joint modeling of multiple objects in a scene naturally handles occlusion through rendering and can reliably track through occlusion.

In Figure~\ref{fig:tracking}(a), the tomato can is briefly occluded by a narrow occluder, and the estimated posterior from particle filtering indicates there is little uncertainty about where the tomato can is even when it is fully occluded. However, accurate uncertainty quantification is important for tracking objects through extended occlusion. To illustrate this point, we generate a synthetic video in which a sugar box moves from left to right and becomes fully occluded by a cracker box. Figure~\ref{fig:tracking}(b) visualizes tracking with 3DNEL with 400 particles in this challenging video. We observe that 3DNEL can accurately quantify uncertainty with particle filtering: the estimated posterior concentrates on the actual pose when the sugar box is visible, yet spreads to cover a range of possible poses when the sugar box becomes occluded. Such modeling of the full posterior helps 3DNEL to regain track when the sugar box reappears, after which the posterior again concentrates on the actual pose. We observe that if we instead use a smaller number of particles (e.g. 50), we would not be able to accurately represent the uncertainty introduced by the occlusion and would lose track.

\subsection{Extension to camera pose tracking from video}\label{sec:tracking}

We demonstrate that 3DNEL's probabilistic formulation provides a principled framework for incorporating prior knowledge about the scene and objects, and enables easy extension to camera pose tracking from video using probabilistic inference in the same model without task specific retraining. We extend our single-frame 3DNEL MSIGP to the multi-timestep setup by introducing a dynamics prior that samples the object pose at time $t+1$ from a Gaussian-VMF distribution restricted to poses with position at most 3cm away from the object pose at time $t$. We initialize object poses at the first frame to ground truth annotations to avoid introducing systematic errors. We again apply stochastic search with 3DNEL, using just ICP and random walk proposals. However, we further assume we know the scene is static and only the camera moves, which translates into jointly updating all object poses by the same amount in a scene. Table \ref{tab:tracking} shows that the same inference procedure can readily handle such extensions, taking into account the dynamics prior and the knowledge of a static scene within the same probabilistic model. We observe comprehensive improvements over single frame predictions.

\subsection{Alternative similarity measurements}\label{sec:dino}

So far our experiments reuse components from SurfEMB to highlight the added benefits of robustness and uncertainty quantification from 3DNEL's principled probabilistic modeling. However, 3DNEL can leverage any learned neural embeddings and similarity measurements. To illustrate this, we demonstrate applying 3DNEL to a simple few-shot pose estimation task in simulation using similarity measurements based on learned neural embeddings from DINOv2~\cite{oquab2023dinov2}, a recently released vision foundation model.

Given 10 training images of a novel object from different perspectives, we reconstruct a 3D mesh of the object using bundle adjustment. For each training image, we sample 2000 pixels on the object surface, and record their corresponding object frame coordinates and DINOv2 embeddings. For a given 3D scene description, we obtain the key embedding of a rendered pixel by finding the closest sampled 3D surface point from the training images and taking the corresponding DINOv2 embedding. We use a VMF distribution with concentration parameter $100.0$ on normalized DINOv2 embeddings as our similarity measurement $\mathbb{P}_{\text{RGB}}$.

We evalaute using the sugar box from YCB-V. We generate a test set consisting of 100 RGB-D images of the sugar box at a fixed position directly in front of the camera, but in randomly sampled orientations. We find that 3DNEL+DINOv2 can accurately infer the object's orientation with an average rotation error of 0.839 degrees. These results suggest a promising future direction to combine the rich, general neural representations from powerful pretrained vision foundation models with the robustness and interpretability of structured probabilistic models.

\section{Conclusion}
In conclusion, we propose a probabilistic inverse graphics approach to pose estimation. We leverage learned neural embeddings and depth information to model likelihood of observed RGB-D images given 3D scene descriptions, and build efficient inference procedures for both pose estimation and tracking. Our approach achieves performance on par with SOTA in sim-to-real setups on YCB-V, and can more robustly handle challenging scenes. Finally, thanks to our probabilistic formulation, we can jointly model all object poses in the scene and easily extend to additional tasks such as uncertainty quantification and camera tracking.

	{\small
		\bibliographystyle{ieee_fullname}
		\bibliography{3dnel}
	}

\end{document}


\title{Supplement to ``3D Neural Embedding Likelihood: Probabilistic Inverse Graphics for Robust 6D Pose Estimation"}

\author{Guangyao Zhou\thanks{Equal contribution} \\
Google DeepMind\\
\texttt{stannis@google.com}
\and
Nishad Gothoskar\footnotemark[1]\\
MIT\\
\texttt{nishad@mit.edu}
\and
Lirui Wang\\
MIT\\
\texttt{liruiw@mit.edu}
\and
Joshua B. Tenenbaum\\
MIT\\
\texttt{jbt@mit.edu}
\and Dan Gutfreund\\
MIT-IBM Watson AI Lab\\
\texttt{dgutfre@us.ibm.com}
\and
Miguel L\'{a}zaro-Gredilla\\
Google DeepMind\\
\texttt{lazarogredilla@google.com}
\and
Dileep George\\
Google DeepMind\\
\texttt{dileepgeorge@google.com}
\and Vikash K. Mansinghka \\
MIT\\
\texttt{vkm@mit.edu}
}

\maketitle
\ificcvfinal\thispagestyle{empty}\fi

\renewcommand\thesection{\Alph{section}}

\section{Review of SurfEMB}\label{sec:surfemb}

Our noise model on RGB information build on SurfEMB \cite{haugaard2022surfemb} which learns neural embeddings to establish dense 2D-3D correspondences via contrastive learning. For each object class $t\in \{1, \cdots, M\}$, SurfEMB learns a pair of neural networks: the \textit{query model} $f_t: \{0, \cdots, 255\}^{H\times W\times 3} \mapsto \mathbb{R}^{H \times W \times E}$ and the \textit{key model} $g_t: \mathbb{R}^{3} \mapsto \mathbb{R}^{E}$ . The \textit{query model} transforms an observed RGB image $\mathbf{I}$ into query embeddings $\mathbf{Q}^t$, while the \textit{key model} transforms a rendered object coordinate image $\mathbf{\tilde{X}}$ into a set of key embeddings.

For a given 2D pixel location on the observed image with query embedding $\mathbf{q}$, SurfEMB specifies a surface correspondence distribution $\mathbb{P}_{\text{RGB}}(g_t(\tilde{x}) | \mathbf{q}, t) \propto \exp(\mathbf{q}^T g_t(\tilde{x}))$ for each object class $t$. To normalize this surface correspondence distribution, for each object class $t$, we subsample uniformly across the object's surface to get a set $Z_t$ of surface 3D coordinates in object frame. Then, given a pixel with query embedding $\mathbf{q}$, we calculate the probability that this pixel corresponds to a surface point $\mathbf{\tilde{x}} \in Z_t$ on object class $t$ as:
\begin{equation}\label{eqn:correspondence}
	\mathbb{P}_{\text{RGB}}(g_t(\mathbf{\tilde{x}}) | \mathbf{q}, Z_t, t) = \frac{\exp(\mathbf{q}^T g_t(\mathbf{\tilde{x}}))}{\sum_{\mathbf{x}\in Z_t} \exp(\mathbf{q}^T g_t(\mathbf{x}))}
\end{equation}

\section{More details on the energy-based formulation of 3DNEL}\label{sec:3dnel}

\subsection{Existence of the normalization constant for the energy-based formulation}
Since we are working with an energy-based formulation (Equation~(1)), to make the probability distribution properly defined we need to make sure the normalization constant, i.e. the sum of the energy function over all $\mathbf{I}$ and $\mathbf{C}$
\begin{equation*}
	\sum_{\mathbf{I}}\int_{\mathbf{C}} \prod_{\mathbf{c}}\left( \epsilon\mathbb{P}_{\text{BG}}(\mathbf{c}; B) + \frac{1 - \epsilon}{\tilde{K}}\sum_{\mathbf{\tilde{c}}:\tilde{s} > 0}
	\mathbb{P}_{\text{depth}}(\mathbf{c} | \mathbf{\tilde{c}}; r)
	\mathbb{P}_{\text{RGB}}(g_{\tilde{s}}(\mathbf{\tilde{x}}) | \mathbf{q}^{\tilde{s}}, \tilde{s}) \right)
\end{equation*}
is finite and well-defined. For RGB images of size $H\times W$, since each pixel has only 256 values, there are at most $256^{H\times W\times 3}$ RGB images of size $H\times W$ which is a finite number. Since the value of the energy function is less than 1 for any given $\mathbf{I}$ and $\mathbf{C}$, summing over a finite number of $\mathbf{I}$ and integrating over a bounded region for $\mathbf{C}$ gives us a finite normalization constant, making the probability distribution well-defined.

\subsection{JAX-based implementation of 3DNEL evaluation given rendering outputs}
Given rendering outputs from OpenGL, we use JAX to develop a 3DNEL evaluation implementation that can run efficiently on modern GPUs. The implementation can be easily combined with \pyobject{jax.vmap} to support 3DNEL evaluation of hundreds of 3D scene descriptions in parallel.
\begin{lstlisting}[language=Python]
# Copyright 2023 DeepMind Technologies Limited
# Copyright 2023 Massachusetts Institute of Technology (M.I.T.)
# SPDX-License-Identifier: Apache-2.0
@functools.partial(jax.jit, static_argnames="filter_shape")
def neural_embedding_likelihood(
    data_xyz: jnp.ndarray,
    query_embeddings: jnp.ndarray,
    log_normalizers: jnp.ndarray,
    model_xyz: jnp.ndarray,
    key_embeddings: jnp.ndarray,
    model_mask: jnp.ndarray,
    obj_ids: jnp.ndarray,
    data_mask: jnp.ndarray,
    r: float,
    p_background: float,
    p_foreground: float,
    filter_shape: Tuple[int, int],
):
    """
    Args:
        data_xyz: Array of shape (H, W, 3). Observed point cloud organized as an image.
        query_embeddings: Array of shape (H, W, n_objs, d).
		Query embeddings for each observed pixel using models from different objects.
        log_normalizers: Array of shape (H, W, n_objs).
		The log normalizers for each pixel given each object model
        model_xyz: Array of shape (H, W, 3). Rendered point cloud organized as an image.
        key_embeddings: Array of shape (H, W, d). Key embeddings organized as an image.
        model_mask: Array of shape (H, W). Mask indicating relevant pixels from rendering.
        obj_ids: Array of shape (H, W). The object id of each pixel.
        data_mask: Array of shape (H, W). Mask indicating the relevant set of pixels.
        r: Radius of the ball.
        p_background: background probability.
        p_foreground: foreground probability.
	  filter_shape: used to restrict likelihood evaluation to a 2D neighborhood.
    """
    obj_ids = jnp.round(obj_ids).astype(jnp.int32)
    padding = [
        (filter_shape[ii] // 2, filter_shape[ii] - filter_shape[ii] // 2 - 1)
        for ii in range(len(filter_shape))
    ]
    model_xyz_padded = jnp.pad(model_xyz, pad_width=padding + [(0, 0)])
    key_embeddings_padded = jnp.pad(key_embeddings, pad_width=padding + [(0, 0)])
    model_mask_padded = jnp.pad(model_mask, pad_width=padding)
    obj_ids_padded = jnp.pad(obj_ids, pad_width=padding)

    @functools.partial(
        jnp.vectorize,
        signature='(m),(n),(o,d),(o)->()',
    )
    def log_likelihood_for_pixel(
        ij: jnp.ndarray,
        data_xyz_for_pixel: jnp.ndarray,
        query_embeddings_for_pixel: jnp.ndarray,
        log_normalizers_for_pixel: jnp.ndarray,
    ):
        """
        Args:
            ij: Array of shape (2,). The i, j index of the pixel.
        """
        model_xyz_patch = jax.lax.dynamic_slice(
            model_xyz_padded,
            jnp.array([ij[0], ij[1], 0]),
            (filter_shape[0], filter_shape[1], 3),
        )
        key_embeddings_patch = jax.lax.dynamic_slice(
            key_embeddings_padded,
            jnp.array([ij[0], ij[1], 0]),
            (filter_shape[0], filter_shape[1], key_embeddings.shape[-1]),
        )
        model_mask_patch = jax.lax.dynamic_slice(model_mask_padded, ij, filter_shape)
        obj_ids_patch = jax.lax.dynamic_slice(obj_ids_padded, ij, filter_shape)
        log_prob_correspondence = (
            jnp.sum(
                query_embeddings_for_pixel[obj_ids_patch] * key_embeddings_patch,
                axis=-1,
            ) - log_normalizers_for_pixel[obj_ids_patch]
        ).ravel()
        distance = jnp.linalg.norm(
            data_xyz_for_pixel - model_xyz_patch, axis=-1
        ).ravel()
        a = jnp.concatenate([jnp.zeros(1), log_prob_correspondence])
        b = jnp.concatenate(
            [
                jnp.array([p_background]),
                jnp.where(
                    jnp.logical_and(distance <= r, model_mask_patch.ravel() > 0),
                    3 * p_foreground / (4 * jnp.pi * r**3),
                    0.0,
                ),
            ]
        )
        log_mixture_prob = logsumexp(a=a, b=b)
        return log_mixture_prob

    log_mixture_prob = log_likelihood_for_pixel(
        jnp.moveaxis(jnp.mgrid[: data_xyz.shape[0], : data_xyz.shape[1]], 0, -1),
        data_xyz,
        query_embeddings,
        log_normalizers,
    )
    return jnp.sum(jnp.where(data_mask, log_mixture_prob, 0.0))
\end{lstlisting}

\section{Details on coarse enumerative pose hypotheses generation}\label{sec:hypotheses}

\subsection{Formal description of the coarse enumerative pose hypotheses generation process}

We develop a novel spherical voting procedure and a heuristic scoring using the query embeddings and observed point cloud image $\mathbf{C}$ defined in Section~3.2, and use them in an enumerative procedure to efficiently generate pose hypotheses. We use the object center and $n_k$ points sampled using farthest point sampling from the object surface as our keypoints, and discretize the camera frame space into a $L_x\times L_y\times L_z$ voxel grid .

\begin{figure}[t!]
	\centering
	\includegraphics[width=\textwidth]{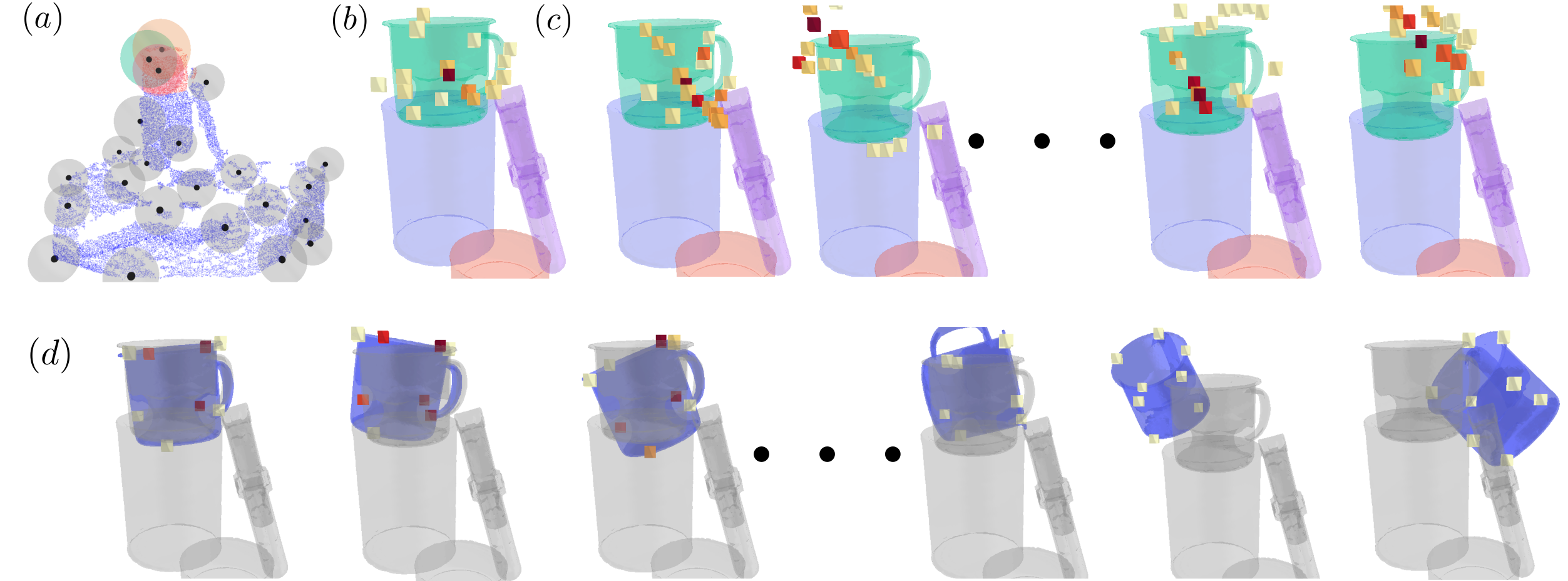}
	\caption{\textbf{Coarse Enumerative Pose Hypotheses Generation} We visualize our coarse enumerative pose hypotheses generation process using a mug object as an example. \textbf{Figure~\ref{fig:voting}(a)} visualizes spherical voting for identifying the mug center. The red points in the observed point cloud represent points associated with the mug. The 3 colored spheres illustrate how votes from points on the mug combine to identify the mug center. The same procedure can also be used to identify the $n_k$ keypoints on the mug surface. \textbf{Figure~\ref{fig:voting}(b)} visualizes 20 top-scoring voxels from the voxel grid associated with the mug center. \textbf{Figure~\ref{fig:voting}(c)} visualizes 20 top-scoring voxels from the voxel grids associated with the $n_k$ keypoints. \textbf{Figure~\ref{fig:voting}(d)} show how we score pose hypotheses by summing the scores of the voxels the corresponding $n_k$ keypoints fall into. The scores of the poses hypotheses decrease as we go from left to right. In \textbf{Figures~\ref{fig:voting}(b) to (d)}, light yellow represents low voxel scores, while dark red represents high voxel scores.}
	\label{fig:voting}
\end{figure}

For a given keypoint, our spherical voting procedure aggregates information from the entire image to score how likely the keypoint is present at different voxel locations, and stores the scores in a voxel grid. Figure~\ref{fig:voting}(a) visualizes spherical voting for the center $x^{*}$ of the mug object: for pixel location $(i, j)$ with camera frame coordinate $\mathbf{c}\in\mathbb{R}^3$ and query embedding $\mathbf{q}\in\mathbb{R}^E$, we identify its most likely corresponding point on the mug surface $x= \argmax_{\tilde{x}} \mathbb{P}_{\text{RGB}}(g_t(\tilde{x})|\mathbf{q}, t)$, calculate the distance $r_x=||x - x^{*}||_2$ from $x$ to $x^{*}$ in the object frame, and cast votes~\cite{qi2019deep} with weight $p_{i,j} = \max_{\tilde{x}}\mathbb{P}_{\text{RGB}}(g_t(\tilde{x})|\mathbf{q}, t)$ towards all points on a sphere of radius $r_x$ centered at $\mathbf{c}$. Figure~\ref{fig:voting}(b) visualizes the 20 top-scoring voxels from the voxel grid for the mug center, and Figure~\ref{fig:voting}(c) visualizes the 20 top-scoring voxels from the voxel grids for the $n_k$ keypoints on the mug surface.

We coarsely discretize the object pose space. We reuse the same camera frame space discretization into a voxel grid, and use the $L_x\times L_y\times L_z$ voxel centers to discretize the location space. We use a customized procedure (Appendix~\ref{sec:rotation_discretization}) to generate $n_r$ rotations and discretize the rotation space. We use the voxel grid for the object center to identify top-scoring object locations, and score all $n_r$ rotations at these locations. We score a given object pose with the sum of the scores of the voxels the corresponding $n_k$ keypoints fall into. Figure \ref{fig:voting}(d) visualizes 3 example top pose hypotheses from the enumerative procedure. Figure \ref{fig:voting}(e) visualizes how poses far away from ground truth get low scores from our heuristic scoring.

In Algorithm \ref{algo:hypotheses} we present a detailed description of our coarse enumerative pose hypotheses generation process. In practice we also optionally apply non-max suppression in the \texttt{TopPositions} function in the algorithm to make sure the promising positions we identify are well spread out to cover different parts of the image and better represent uncertainty.

\begin{algorithm}[h!]
	\caption{Coarse Enumerative Pose Hypotheses Generation}
	\label{algo:hypotheses}
	\RestyleAlgo{ruled}
	\SetKwProg{Def}{def}{}{end}
	\SetKwFunction{vote}{Voting}
	\SetKwFunction{position}{TopPositions}
	\SetKwFunction{score}{Scoring}
	\SetKwFunction{rank}{RankByScore}
	\SetKwInput{Object}{Object-specific}
	\SetKwInput{Spatial}{Spatial discretization}
	\SetKwInput{Orientation}{Orientation discretization}
	\SetKwInput{Parameters}{Parameters}
	\tcc{Basic setups}
	\Object{A set of surface points $Z_t$ for object class $t$, query model $f_t$, key model $g_t$, $n_k$ keypoints with object frame coordinates $x^{*}_1, \cdots, x^{*}_{n_k} \in \mathbb{R}^3$.}
	\Spatial{
		Camera frame coordinates of the center of the boundary voxel $y\in \mathbb{R}^3$, size of the voxel grid $(L_x, L_y, L_z)$, diameter of the voxels $d > 0$.
	}
	\Orientation{$n_r$ representative orientations $\mathbf{R}_1, \cdots, \mathbf{R}_{n_r} \in \mathbb{SO}(3).$}
	\Parameters{Number of pose hypotheses to generate $n_p$, number of top positions $n_t$.}
	\hrulefill

	\tcc{Coarse Enumerative pose hypotheses generation}
	\KwIn{RGB image $\mathbf{I}\in\mathbb{R}^{H\times W\times 3}$, observed point cloud $\mathbf{C}\in\mathbb{R}^{H\times W\times 3}$.}
	\KwOut{Top scoring pose hypotheses $\mathbf{P}^t_1, \cdots, \mathbf{P}^t_{n_p} \in \mathbb{SE}(3)$.}
	$\mathbf{Q} \leftarrow f_t(\mathbf{I})$\tcp*{Get query embeddings $\mathbf{Q}\in \mathbb{R}^{H\times W\times E}$ from the RGB image $\mathbf{I}$}
	$\mathbf{V}^0 \leftarrow$ \vote($\mathbf{Q}, \mathbf{C}, (0, 0, 0)$)\tcp*{Aggregation for object center $(0, 0, 0)$}
	\For(\tcp*[f]{Aggregation for $n_k$ keypoints.}){$i \leftarrow 1$ \KwTo $n_k$}{
		$\mathbf{V}^i \leftarrow$ \vote($\mathbf{Q}, \mathbf{C}, x^{*}_i$)\tcp*{$\mathbf{V}^i \in \mathbb{R}^{L_x\times L_y\times L_z}$}
	}
	\tcc{Identify top positions based on $\mathbf{V}^0$'s largest entires.}
	$l_1, \cdots, l_{n_t}\leftarrow $ \position($\mathbf{V}^0$)\tcp*{$l_1, \cdots, l_{n_t}\in \mathbb{R}^3$}
	\For{$i \leftarrow 1$ \KwTo $n_t$, $j \leftarrow 1$ \KwTo $n_r$}{
		$s_{i,j} \leftarrow$ \score($l_i, \mathbf{R}_j, \mathbf{V}^1, \cdots, \mathbf{V}^{n_k}$)\tcp*{Heuristic pose scoring}
	}
	$\mathbf{P}^t_1, \cdots, \mathbf{P}^t_{n_p} \leftarrow $ \rank($l_1, \cdots, l_{n_t}, s_{i,j}, i=1, \cdots, n_t, j=1, \cdots, n_r$)\;
	\KwRet{$\mathbf{P}^t_1, \cdots, \mathbf{P}^t_{n_p}$}\;
	\hrulefill

	\tcc{Voting and heuristic scoring}
	\Def{\vote($\mathbf{Q}, \mathbf{C}, x^{*}$)\tcp*[f]{Voting-based evidence aggregation}}{
		$\mathbf{V} \leftarrow \mathbf{0}$\tcp*{Initialize $V \in \mathbb{R}^{L_x\times L_y\times L_z}$ to all $0$ array}
		\For{$i\leftarrow 1$ \KwTo $H$, $j\leftarrow 1$ \KwTo $W$}{
			$x\leftarrow \argmax_{\tilde{x}} \mathbb{P}_{RGB}(\tilde{x}|Q_{i, j}, Z_t, t)$, $p_{i,j} \leftarrow \max_{\tilde{x}}\mathbb{P}_{RGB}(\tilde{x}|Q_{i, j}, Z_t, t)$\;
			\For{$u\leftarrow 1$ \KwTo $L_x$, $v\leftarrow 1$ \KwTo $L_y$, $w\leftarrow 1$ \KwTo $L_z$}{
				$c \leftarrow \left(y_1 + (u-1)d, y_2 + (v - 1)d, y_3 + (w-1)d\right)$\;
				\uIf{$||\mathbf{C}_{i,j} - c||_2 \approx || x - x^{*}||_2 $}{$\mathbf{V}_{u,v,w} = \mathbf{V}_{u,v,w} + p_{i,j}$}
			}
		}
		\KwRet{$\mathbf{V}$}
	}
	\Def{\score($l, \mathbf{R}, \mathbf{V}^1, \cdots, \mathbf{V}^{n_k}$)\tcp*[f]{Heuristic pose scoring}}{
	$s \leftarrow 0$\tcp*{Initialize score to 0}
	\For{$i \leftarrow 1$ \KwTo $n_k$}{
	$x \leftarrow \mathbf{R}x^{*}_i + l$\tcp*{Location of $x^{*}_i$ in world frame for pose $l, \mathbf{R}$}
	$(u, v, w) \leftarrow Round[(x-y)/d]$\tcp*{Identify corresponding voxel of $x$}
	$s = s + \mathbf{V}^i_{u,v,w}$\tcp*{Heuristic scoring}
	}
	\KwRet{$s$}
	}
\end{algorithm}

\subsection{Taichi-based spherical voting}
We use Taichi~\cite{hu2019taichi} to develop a spherical voting implementation that can run efficiently on modern GPUs.
\begin{lstlisting}[language=Python]
# Copyright 2023 DeepMind Technologies Limited
# Copyright 2023 Massachusetts Institute of Technology (M.I.T.)
# SPDX-License-Identifier: Apache-2.0

@ti.kernel
def taichi_spherical_vote(
    centers: ti.types.ndarray(element_dim=1),
    radiuses: ti.types.ndarray(),
    weights: ti.types.ndarray(),
    voxel_grid: ti.types.ndarray(),
    voxel_grid_start: ti.types.ndarray(element_dim=1),
    voxel_diameter: float,
    multipliers: ti.types.ndarray(),
):
    """
    Args:
        centers: Array of shape (batch_size, n_centers, 3,). Coordinates of the centers of the spheres.
        radiusss: Array of shape (batch_size, n_centers). Radiuses of the spheres.
        weights: Array of shape (batch_size, n_centers,). Weights of votes from the spheres.
        voxel_grid: Array of shape voxel_grid_shape
        voxel_grid_start: Array of shape (3,). Coordinate of the center of voxel (0, 0, 0)
        voxel_diameter: float. Diameter of a voxel.
	  multipliers: fixed length-2 1D array with elements 1.0, -1.0
    """
    for voxel in ti.grouped(voxel_grid):
        voxel_grid[voxel] = 0.0

    for ii, jj in centers:
        center_on_voxel_grid = (
            centers[ii, jj] - voxel_grid_start[None]
        ) / voxel_diameter
        center_on_voxel_grid = ti.round(center_on_voxel_grid)
        radius_in_voxels = radiuses[ii, jj] / voxel_diameter + 0.5
        for x in range(ti.ceil(radius_in_voxels)):
            for y in range(ti.ceil(ti.sqrt(radius_in_voxels**2 - x**2))):
                z_range = (
                    ti.ceil(
                        ti.sqrt(
                            ti.max(
                                0.0,
                                (radiuses[ii, jj] / voxel_diameter - 0.5) ** 2
                                - x**2
                                - y**2,
                            )
                        )
                    ),
                    ti.ceil(ti.sqrt(radius_in_voxels**2 - x**2 - y**2)),
                )
                for z in range(z_range[0], z_range[1]):
                    for xx in range(2):
                        if x == 0 and multipliers[xx] < 0:
                            continue

                        x_coord = ti.cast(
                            center_on_voxel_grid[0] + multipliers[xx] * x,
                            ti.i32,
                        )
                        if x_coord < 0 or x_coord >= voxel_grid.shape[1]:
                            continue

                        for yy in range(2):
                            if y == 0 and multipliers[yy] < 0:
                                continue

                            y_coord = ti.cast(
                                center_on_voxel_grid[1] + multipliers[yy] * y,
                                ti.i32,
                            )
                            if y_coord < 0 or y_coord >= voxel_grid.shape[2]:
                                continue

                            for zz in range(2):
                                if z == 0 and multipliers[zz] < 0:
                                    continue

                                z_coord = ti.cast(
                                    center_on_voxel_grid[2] + multipliers[zz] * z,
                                    ti.i32,
                                )
                                if z_coord < 0 or z_coord >= voxel_grid.shape[3]:
                                    continue

                                ti.atomic_add(
                                    voxel_grid[ii, x_coord, y_coord, z_coord],
                                    weights[ii, jj],
                                )
\end{lstlisting}

\subsection{Discretizing the rotation space}\label{sec:rotation_discretization}

We discretize $\mathbb{SO}(3)$ into 6400 representative orientations. We generate these orientations by first picking 200 points roughly uniformly on the unit sphere using the Fibonacci sphere. The 6400 representative orientations are genearted by first rotating the axis $(0, 0, 1.0)$ to point to one of the 200 points, followed by one of $32$ in-plane rotations around the axis.

\section{Details on inference pipeline implementation}\label{sec:inference_implementation}

\subsection{Using additional 2D detection and mask predcition from SurfEMB}

In the best performing setup, we leverage the same 2D detector used in SurfEMB as part of the pose hypotheses generation process. Although our spherical voting procedure can robustly aggregate information from the entire image to generate pose hypotheses, as demonstrated by the competitive performance of \textit{3DNEL MSIGP (No 2D detection)} (Abalations in Table~1), in practice the query embedding images for many objects are very noisy and tend to hurt performance. Empirically, we observe that by additionally using 2D detections, we can focus the spherical voting process on regions of the observed image that is likely relevant for the objects, and further improve performance.

For each object class in the scene, there can be multiple 2D detections. For each 2D detection, we do spherical voting just within the detector crop and generate 80 pose hypotheses per detector crop. When there is a missing 2D detection, we obtain the query embeddings by upsampling the input RGB image by 1.5x, and do spherical voting on the whole image. In such cases, when we identify top positions, we additionally do non-max suppression with a filter size of $10$ to spread the top-scoring positions out. For each such top-scoring position we identify top 2 orientation, and we consider all top-scoring positions and generate in total $30$ pose hypotheses.

\begin{figure}[t!]
	\centering
	\includegraphics[width=\linewidth]{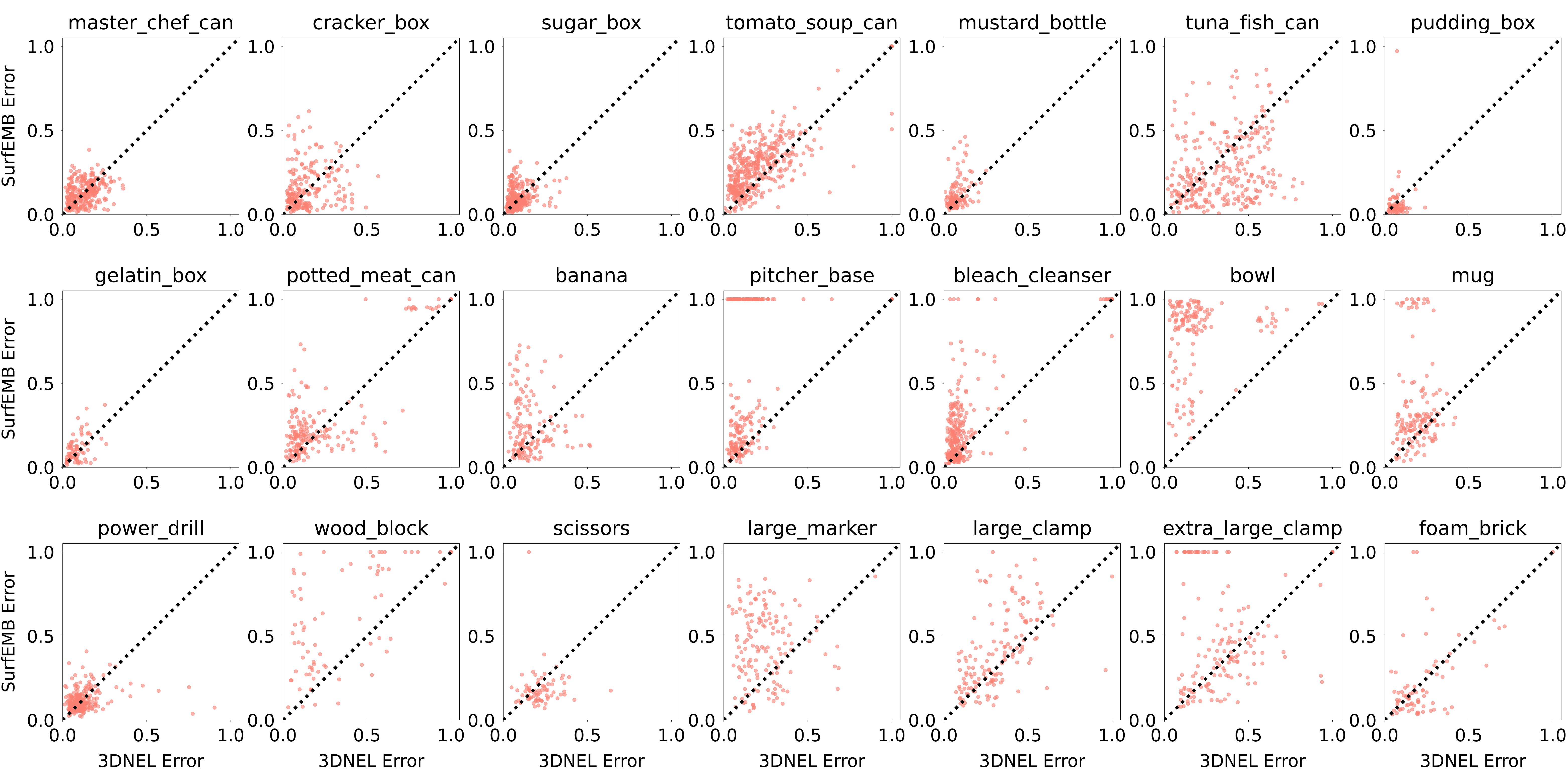}
	\caption{We compare the prediction error (using the VSD error metric) of SurfEMB and 3DNEL MSIGP across all 4123 object instances in the YCB-V test dataset, with each instance represented as a point on this scatter plot. In the main text Figure~3(a), we show the scatter plot across all objects. Here, we show the results per object.}
	\label{fig:full_scatter_plot}
\end{figure}

\subsection{Implementation details and hyperparameters}

We use OpenGL for rendering, use Taichi for spherical voting, and use JAX for 3DNEL evaluation. We refer the readers to the attached Python source code for a complete implementation of our pipeline.

As we describe in the main text, we pick hyperparameters by visually inspecting detection results on a small number of real training images that are outside the test set.

\begin{figure}[t!]
	\centering
	\includegraphics[width=0.65\linewidth]{images/qualitative_examples_all.pdf}
	\caption{Additional visualizations of SurfEMB and 3DNEL MSIGP predictions on YCB-V test images}
	\label{fig:more_qualitative_examples}
\end{figure}

We select $n_k = 8$ keypoints from the surface of each object class, and use $y=(-350.0, -210.0, 530.0), L_x=129, L_y=87, L_z=168$ and $d = 5.0$ to make the voxel grids large enough to cover all the keypoints that can be present in the camera frame. Here the units for the values in $y$ and $d$ are mm.

We use $r=5.0$ in evaluating 3DNEL for all our experiments.  When identifying points in the rendered point cloud $\mathbf{\tilde{C}}$ (organized as an $H\times W\times 3$ image) that is within distance $r$ from a point $(i, j)$ in the observed point cloud $\mathbf{C}$, to further speed up 3DNEL evaluation, we only look at the points from the patch of size $(10, 10)$ centered at $(i, j)$.

In Figure \ref{fig:more_qualitative_examples} we include additional visualizations of SurfEMB and 3DNEL MSIGP predictions on YCB-V test images.

\section{Additional robustness results}
In Figure \ref{fig:full_scatter_plot} we include additional robustness results.

\section{Additional visualizations of SurfEMB and 3DNEL MSIGP predictions on YCB-V test images}

 {\small
  \bibliographystyle{ieee_fullname}
  \bibliography{3dnel}
 }

\newpage
\renewcommand{\thefigure}{A\arabic{figure}}

\setcounter{figure}{0}